\documentclass[graybox]{svmult}

\usepackage{type1cm}        
\usepackage{makeidx}         
\usepackage{graphicx}        
\usepackage{multicol}        
\usepackage[bottom]{footmisc}

\usepackage{newtxtext}       %
\usepackage{newtxmath}       

\usepackage{algorithm}
\usepackage{algorithmic}
\usepackage{subcaption}
\makeindex             

\newcommand{\norm}[1]{\left\Vert#1\right\Vert}

\DeclareMathOperator{\argmin}{arg\,min}

\DeclareMathAlphabet{\pazocal}{OMS}{zplm}{m}{n}

\newcommand{\mb}[1]{\mathbf{#1}}
\newcommand{\bx}[0]{\mb{x}}
\newcommand{\bX}[0]{\mb{X}}
\newcommand{\bz}[0]{\mb{z}}
\newcommand{\bI}[0]{\mb{I}}
\newcommand{\tbz}[0]{\tilde{\mb{z}}}
\newcommand{\btheta}[0]{\mb{\theta}}
\newcommand{\bphi}[0]{\mb{\phi}}

\newcommand{\Ex}[2]{\mathbb{E}_{{#1}}\Big[ {#2} \Big]}
\newcommand{\mathL}[0]{\mathcal{L}}
\newcommand{\tmathL}[0]{\tilde{\mathcal{L}}}

\begin{document}

\title*{Autoencoders}
\author{Dor Bank, Noam Koenigstein, Raja Giryes}
\institute{ Dor Bank \at School of Electrical Engineering, Tel Aviv University, \email{dorbank@mail.tau.ac.il}
\and Noam Koenigstein \at Department of Industrial Engineering, Faculty of Engineering, Tel Aviv University, \email{noamk@tauex.tau.ac.il}
\and Raja Giryes \at School of Electrical Engineering, Tel Aviv University, \email{raja@tauex.tau.ac.il}}
%
%
\maketitle

\abstract{
An autoencoder is a specific type of a neural network, which is mainly designed to encode the input into a compressed and meaningful representation, and then decode it back such that the reconstructed input is similar as possible to the original one. This chapter surveys the different types of autoencoders that are mainly used today. It also describes various applications and use-cases of autoencoders.}

\section{Autoencoders}
\label{sec:autoencoders}
\begin{figure}
    \centering
    \includegraphics[scale=0.3]{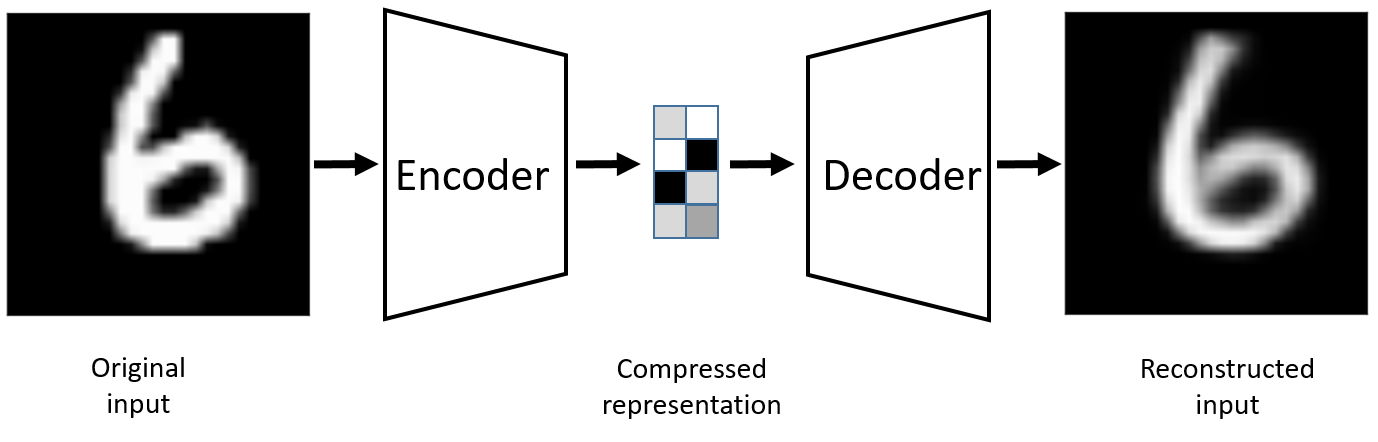}
    \caption{An autoencoder example. The input image is encoded to a compressed representation and then decoded.}
    \label{fig:Autoencoder}
\end{figure}

Autoencoders have been first introduced in \cite{AutoEncoder_original} as a neural network that is trained to reconstruct its input. Their main purpose is learning in an unsupervised manner an ``informative'' representation of the data that can be used for various implications such as clustering. The problem, as formally defined in \cite{AutoEncoders_explanation}, is to learn the functions $A: \mathbb{R}^n \rightarrow \mathbb{R}^p$ (encoder) and $B: \mathbb{R}^p \rightarrow \mathbb{R}^n$ (decoder) that satisfy
\begin{equation}
\argmin_{A,B}E[\Delta(\bx, B\circ A (\bx)],
\end{equation}
where $E$ is the expectation over the distribution of $x$, and $\Delta$ is the reconstruction loss function, which measures the distance between the output of the decoder and the intput. The latter is usually set to be the $\ell_2$-norm. Figure \ref{fig:Autoencoder} provides an illustration of the autoencoder model.

In the most popular form of autoencoders, $A$ and $B$ are neural networks \cite{NNAutoEncoder}.
In the special case that $A$ and $B$ are linear operations, we get a linear autoencoder \cite{linear_AutoEncoders}. In the case of linear autoencoder where we also drop the non-linear operations, the autoencoder would achieve the same latent representation as Principal Component Analysis (PCA) \cite{PCA_linearautoencoder}. Therefore, an autoencoder is in fact a generalization of PCA, where instead of finding a low dimensional hyperplane in which the data lies, it is able to learn a non-linear manifold.

Autoencoders may be trained end-to-end or gradually layer by layer. In the latter case, they are ''stacked'' together, which leads to a deeper encoder. In \cite{ConvAutoEncoder}, this is done with convolutional autoencoders, and in \cite{Stacked_autoEncoders} with denoising autoencoder (described below).

This chapter is organized as follows. In Section~\ref{sec:Regularized_autoencoders}, different regularization techniques for autoencoders are considered, whose goal is to ensure that the learned compressed representation is meaningful. In Section~\ref{sec:variational_autoencoders}, the variational autoencoders are presented, which are considered to be the most popular form of autoencoders. Section~\ref{sec:Applications_of_autoencoders} covers very common applications for autoencoders, 
Section~\ref{sec:autoencoders_and_generative_adversarial_networks} briefly discusses the comparison between autoencoders and generative adversarial networks, 
and Section~\ref{sec:Advanced_autoencoder_techniques} describes some recent advanced techniques in this field. Section~\ref{sec:Conclusions and open problems} concludes this chapter.

\section{Regularized autoencoders}
\label{sec:Regularized_autoencoders}
Since in training, one may just get the identity operator for $A$ and $B$, which keeps the achieved representation the same as the input, some additional regularization is required. The most common option is to make the dimension of the representation smaller than the input. This way, a $bottleneck$ is imposed. This option also directly serves the goal of getting a low dimensional representation of the data. This representation can be used for purposes such as data compression, feature extraction, etc. Its important to note that even if the $bottleneck$ is comprised of only one node, then overfitting is still possible if the capacity of the encoder and the decoder is large enough to encode each sample to an index.

In cases where the size of the hidden layer is equal or greater than the size of the input, there is a risk that the encoder will simply learn the identity function. To prevent it without creating a bottleneck (i.e. smaller hidden layer) several options exists for regularization, which we describe hereafter, that would enforce the autoencoder to learn a different representation of the input.

An important tradeoff in autoencoders is the bias-variance tradeoff. On the one hand, we want the architecure of the autoencoder to be able to reconstruct the input well (i.e. reduce the reconstruction error). On the other hand, we want the low representation to generalize to a meaningful one.
We now turn to describe different methods to tackle such tradeoffs.

\subsection{Sparse Autoencoders}
\label{sec:sparse_autoencoders}
One way to deal with this tradeoff is to enforce sparsity on the hidden activations. This can be added on top of the bottleneck enforcement, or instead of it. There are two strategies to enforce the sparsity regularization. They are similar to ordinary regularization, where they are applied on the activations instead of the weights.
The first way to do so, is to apply $L_1$ regularization, which is known to induce sparsity.  Thus, the autoencoder optimization objective becomes:
\begin{equation}
\argmin_{A,B}E[\Delta(\bx, B\circ A (\bx)]+\lambda\sum_i{|a_i|},
\end{equation}
where $a_i$ is the activation at the $i$th hidden layer and $i$ iterates over all the hiddens activations.
Another way to do so, is to use the KL-divergence, which is a measure of the distance between two probability distributions. Instead of tweaking the $lambda$ parameter as in the $L_1$ regularization, we can assume the activation of each neuron acts as a Bernouli variable with probability $p$ and tweak that probability. At each batch, the actual probability is then measured, and the difference is calculated and applied as a regularization factor. For each neuron $j$, the calculated empirical probability is $\hat{p}_j=\frac{1}{m}\sum_i{a_i(x)}$, where $i$ iterates over the samples in the batch.
Thus the overall loss function would be
\begin{equation}
\argmin_{A,B}E[\Delta(\bx, B\circ A (\bx)]+\sum_j{KL(p||\hat{p}_j)},
\end{equation}
where the regularization term in it aims at matching $p$ to $
\hat{p}$.

\subsection{Denoising Autoencoders}
\label{sec:denoising_autoencoders}
\begin{figure}
    \centering
    \includegraphics[scale=0.8]{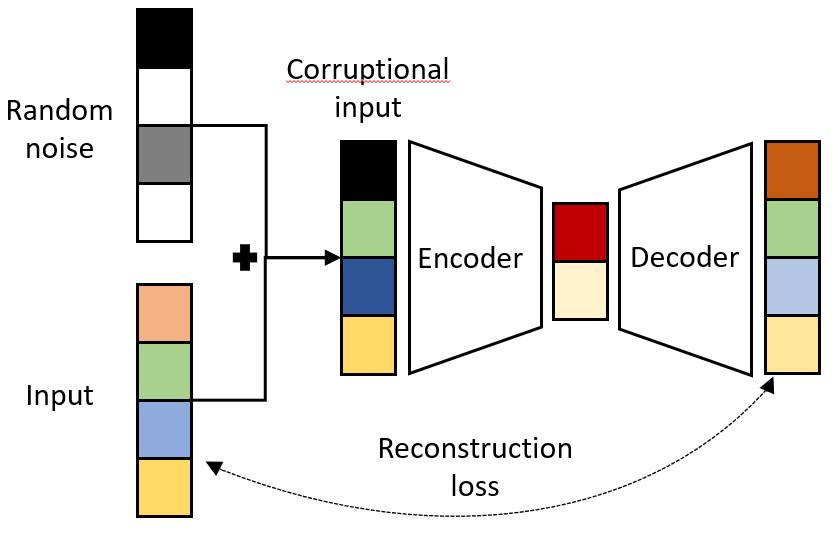}
    \caption{A denoising autoencoder example. The disrupted input image is encoded to a representation and then decoded.}
    \label{fig:Denoised_Autoencoder}
\end{figure}

Denoising autoencoders \cite{Denoising_AutoEncoders} can be viewed either as a regularization option, or as robust autoencoders which can be used for error correction. In these architectures, the input is disrupted by some noise (e.g., additive white Gaussian noise or erasures using Dropout) and the autoencoder is expected to reconstruct the clean version of the input, as illustrated in Figure \ref{fig:Denoised_Autoencoder}.

Note that $\Tilde{\bx}$ is a random variable, whose distribution is given by $C(\Tilde{\bx}|\bx)$. Two common options for $C$ are:
\begin{equation}
C_\sigma(\Tilde{\bx}|\bx) = \mathcal{N}(\bx,\sigma^2 \mathcal{I}),
\end{equation}
and
\begin{equation}
C_p(\Tilde{\bx}|\bx) = \beta \odot \bx, ~~ \beta\sim Ber(p),
\end{equation}
where $\odot$ detnotes an element-wise (Hadamard) product.
In the first option, the variance parameter $\sigma$  sets the impact of the noise. In the second, the parameter $p$ sets the probability of a value in $\bx$ not being nullified.
A relationship between denoising autoencoders with dropout to analog coding with erasures has been shown in \cite{denoising_ae_and_ETF}.

\subsection{Contractive Autoencoders}
\label{sec:contractive_autoencoders}
In denoising autoencoders, the emphasis is on letting the encoder be resistant to some perturbations of the input. In contractive autoencoders, the emphasis is on making the feature extraction less sensitive to small perturbations, by forcing the encoder to disregard changes in the input that are not important for the reconstruction by the decoder.
Thus, a penatly is imposed on the Jacobian of the network. 
The Jacobian matrix of the hidden layer $h$ consists of the derivative of each node $h_j$ with respect to each value $x_i$ in the input $x$. Formally: $J_{ji}=\nabla_{x_i}h_j(x_i)$. In contractive autoencoders we try to minimize its L2 norm, such that the overall optimization loss would be:
\begin{equation}
\argmin_{A,B}E[\Delta(x, B\circ A (x)]+\lambda||J_A(x)||^2_2.
\end{equation}

The reconstruction loss function and the regularization loss actually pull the result towards opposite directions. By minimizing the squared Jacobian norm, all the latent representations of the input tend to be more similar to each other, and by thus make the reconstruction more difficult, since the differences between the representations are smaller. The main idea is that variations in the latent representation that are not important for the reconstructions would be diminished by the regularization factor, while important variations would remain because of their impact on the reconstruction error.

\section{Variational Autoencoders}
\label{sec:variational_autoencoders}
\begin{figure}
    \centering
    \includegraphics[scale=0.5]{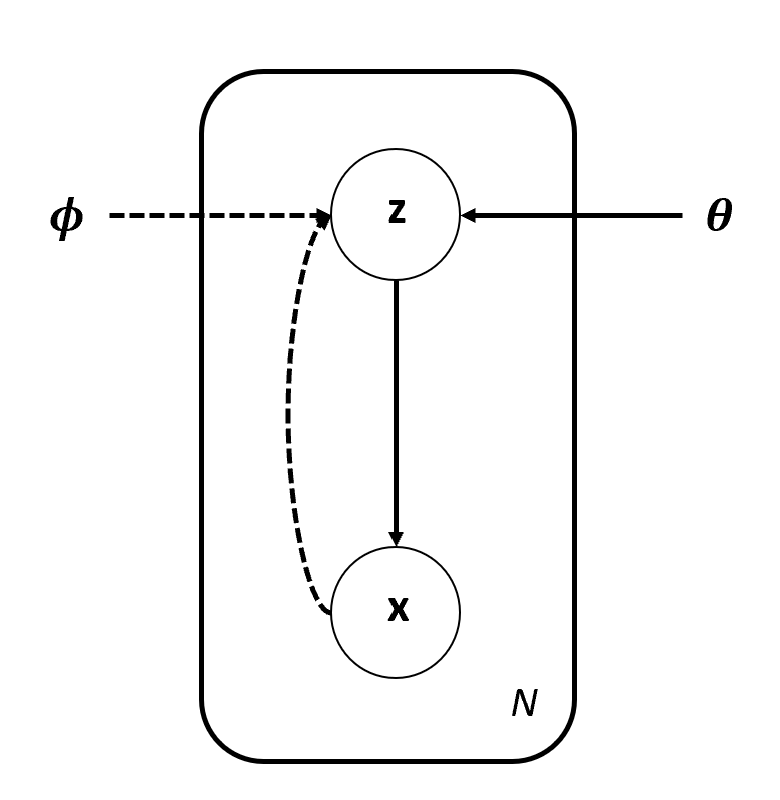}
    \caption{A Graphical Representatin of VAE}
    \label{fig:generativeModel}
\end{figure}

A major improvement in the representation capabilities of autoencoders has been achieved by the Variational Autoencoders (VAE) model \cite{VariationalAutoEncoder}. 
Following Variational Bayes (VB) Inference \cite{BishopBook}, VAE are generative models that attempt to describe data generation through a probabilistic distribution. 
Specifically, given an observed dataset $\mb{X}=\{\bx_i\}_{i=1}^N$ of $V$ i.i.d samples, 
we assume a generative model for each datum $\bx_i$ conditioned on an unobserved random latent variable $\bz_i$, where $\btheta$ are the parameters governing the generative distribution. This generative model is also equivalent to a \emph{probabilistic decoder}. Symmetrically, we assume an approximate posterior distribution over the latent variable $\bz_i$ given a datum $\bx_i$ denoted by recognition,  which is equivalent a \emph{probabilistic encoder} and governed by the parameters $\bphi$. Finally, we assume a prior distribution for the latent variables $\bz_i$ denoted by $p_\btheta \left( \bz_i \right)$. 
Figure \ref{fig:generativeModel} depicts the relationship described above. The parameters $\btheta$ and $\bphi$ are unknown and needs to learned from the data. The observed latent variables $\bz_i$ can be interpreted as a \emph{code} given by the \emph{recognition model}  $q_{\bphi} \left( \bz | \bx \right)$.

The marginal log-likelihood is expressed as a sum over the individual data points $\log p_\btheta (\bx_1, \bx_2,...,\bx_N)=\sum_{i=1}^N \log p_\btheta(\bx_i)$, and each point can be rewritten as: 
\begin{equation}
\log{p_\btheta(\bx_i)}= D_{KL}\left( q_{\bphi} \left( \bz | \bx_i \right) || p_{\btheta} (\bz|\bx_i) \right) +\mathcal{L}(\btheta,\bphi ; \bx_i),
\end{equation}
where the first term is the Kullback-Leibler divergence of the approximate recognition model from the true posterior and the second term is called the \emph{variational lower bound} on the marginal likelihood defined as:
\begin{equation}
\label{eq:VLB_Defined}
    \mathcal{L}(\btheta,\bphi ; \bx_i) \triangleq \Ex{q_\bphi (\bz | \bx_i)}{ -\log{q_\bphi (\bz|\bx)} + \log{p_\btheta (\bx,\bz)} } .
\end{equation}
Since the Kullback-Leibler divergence is non-negative, $\mathcal{L}(\btheta,\bphi ; \bx_i)$ is a lower bound on the marginal log-likelihood and the marginal log-likelihood is independent of the parameters $\btheta$ and $\bphi$, maximizing the lower bound improves our approximation of the posterior with respect to the Kullback-Leibler divergence. 

The variational lower bound can be further expanded as follows: 
\begin{equation}
\label{eq:variational_lower_bound}
    \mathL(\btheta,\bphi ; \bx_i) = -D_{KL} \left( q_\bphi (\bz|\bx_i) || p_\btheta (\bz) \right) + \Ex{q_\bphi(\bz|\bx_i)}{\log{p_\btheta(\bx_i|\bz)}}
\end{equation}
Variational inference follows by maximizing $\mathL(\btheta,\bphi ; \bx_i)$ for all data points with respect to $\btheta$ and $\bphi$.

Given a dataset $\mb{X}=\{\bx_i\}_{i=1}^N$ with $N$ data points, we can estimate the marginal likelihood lower-bound of the full dataset $\mathcal{L} ( \btheta,\bphi ; \bX )$ using a mini-batch $\bX^M = \left\{ \bx_i \right\}_{i=1}^M$ of size $M$ as follows:
\begin{equation}
\label{eq:VLB_on_X}
    \mathL(\btheta,\bphi ; \bX) \approx \tmathL^M (\btheta,\bphi ; \bX^M) = \frac{N}{M}\sum_{i=1}^M\mathL(\btheta,\bphi ; \bx_i)
\end{equation}
Classical mean-field VB assumes a factorized approximate posterior followed by a closed form optimization updates (which usually required conjugate priors). However, VAE follows a different path in which the gradients of $\tilde{\mathcal{L}}^M (\btheta,\bphi ; \bX^M)$ are approximated using a the reparameterization trick and stochastic gradient optimization.

\subsection{The Reparameterization Trick}
The reparameterization trick is a simple approach to  estimate 
$\mathL(\btheta,\bphi ; \bx_i)$ based on a small sample of size $L$. Consider Equation \ref{eq:VLB_Defined}, we can reparameterize the random variable $\tbz \sim q_\bphi(\bz|\bx)$ using a differentiable transformation 
$g_\bphi(\epsilon,\bx)$ using an auxiliary noise vrabile $\epsilon$ drawn from some distribution $\epsilon \sim p(\epsilon)$ \cite{VariationalAutoEncoder}.
Using this tecnique, $\mathL(\btheta,\bphi ; \bx_i)$ is approximated as follows:
\begin{equation}
\label{eq:reparameterization}
    \mathL(\btheta,\bphi ; \bx_i) \approx \tmathL(\btheta,\bphi ; \bx_i) =
    \frac{1}{L} \sum_{l=1}^L \log{p_\btheta(\bx_i,\bz_{(i,l)})}-\log{q_{\bphi}(\bz_{(i,l)|\bx_i})},
\end{equation}
where $\bz_{(i,l)}=g_\bphi(\epsilon_{(i,l)},\bx_i)$ and $\epsilon_{(i,l)}$ is a random noise drawn from $\epsilon_l \sim p(\epsilon)$.


Remember we wish to optimize the mini-batch estimates from Equation \ref{eq:VLB_on_X}. By plugging Equation \ref{eq:reparameterization} we get the following differentiable expression: 
\begin{equation}
    \label{eq:hatMathLDefined}
    \hat{\mathL}^{M} (\btheta,\bphi ; \bX) = \frac{N}{M}\sum_{i=1}^M\tmathL(\btheta,\bphi ; \bx_i),
\end{equation}
which can be derived according to $\btheta$ and $\bphi$ and plugged into an optimizer framework. 

\begin{algorithm}
\caption{Pseudo-code for VAE}
\begin{algorithmic}
\STATE $ (\btheta, \bphi) \gets$ Initialize Parameter
\REPEAT
\STATE $\bX^M \gets$ Random minibatch of $M$ datapoints
\STATE $\epsilon \gets$ $L$ random samples of $p(\epsilon)$ 
\STATE $\bf{g} \gets \nabla_{(\btheta, \bphi)}\hat{\mathL}^{M} (\btheta,\bphi ; \bX)$
\COMMENT{Gradients of Equation \ref{eq:hatMathLDefined}}
\STATE $ (\btheta, \bphi) \gets$ Update parameters based on $\bf{g}$
\COMMENT{e.g., update with SGD or Adagrad}
\UNTIL{Convergenge of $ (\btheta, \bphi)$}
\RETURN $ (\btheta, \bphi)$
\end{algorithmic}
\label{alg:VAE}
\end{algorithm}

Algorithm \ref{alg:VAE} summarizes the full optimization procedure for VAE. Often $L$ can be set to $1$ so long as $M$ is large enough. Typical numbers are $M=100$ and $L=1$.

Equation\ref{eq:reparameterization} presents a lower bound on the log-likelihood $\log{p_\btheta(\bx_i)}$. In \cite{IWAE}, the equation is changed to 
\begin{equation}
\label{eq:IWAE}
    \mathL(\btheta,\bphi ; \bx_i) =
    \frac{1}{L} \sum_{l=1}^L  \log{\frac{1}{k} \sum_{j=1}^k\frac{p_\btheta(\bx_i,\bz_{(j,l)})}{q_{\bphi}(\bz_{(j,l)|\bx_i})}}.
\end{equation}

Intuitively, instead of taking the gradient of a single randomized latent representation, the gradients of the generative network are learned by a weighted average of the sample over different samples from its (approximated) posterior distribution. The weights simply the likelihood functions $q_{\bphi}(\bz_{(j,l)|\bx_i})$.

\subsection{Example: The Case of Normal Distribution}
Usually, we approximate $p(\bz|\bx)$ with a Gaussian distribution $q_\bphi (\bz|\bx)=\mathcal{N}(g(\bx),h(\bx))$, where $g(\bx)$ and $h(\bx)$ are the mean and the covariance of the distribution defined by the encoder network. Namely, the encoder takes an input $\bx_i$ and maps it into a mean and covariance that determine the approximate posterior distribution $q_\bphi (\bz|\bx)$. 

To enable backpropagation through the network, sampling from $q_\bphi (\bz|\bx)$ can simplified using the reparametrisation trick as follows:
\begin{equation}
\bz = h(\bx)\xi + g(\bx), 
\end{equation}
where $\xi \sim \mathcal{N}(0,\bI)$ is a normal distribution. 

Finally, we denote the decoder with an additional function $f$, and require that $x \approx f(\bz)$. The loss function of the entire network then becomes:
\begin{equation}
    loss = c \norm{x-f(\bz)}^2+ D_{KL}\left(\mathcal{N}(g(\bx), h(\bx)), \mathcal{N}(0,\bI) \right),
\end{equation}
which can be automatically derived with respect to the network parameters in $g,h$ and $f$ and optimized with backpropogation.

\subsection{Disentangled Autoencoders}
\label{sec:disentangled_autoencoders}
The variational lower bound as presented at Eq.  \ref{eq:variational_lower_bound}, can be viewed as the summation of two terms: The right term that includes the reconstruction capability of samples, and the left term that acts as a regularization that biases $q_\phi(z|\bx^{(i)}$ towards the assumed prior $p_\theta(z)$.
Disentangled autoencoders include variational autoencoders with a small addition. They add a parameter $\beta$ is as a multiplicative factor for the $KL$ divergence \cite{betaVAELB} at Eq. \ref{eq:variational_lower_bound}. Its maximization factor is thus:
\begin{equation}
\mathL(\theta,\phi,\bx^{(i)})=-\beta D_{KL}(q_\phi(z|\bx^{(i)})||p_\theta(z))+\mathbb{E}_{q_\phi(z|\bx^{(i)})}[\log{p_\theta(\bx^{(i)}|z)}].
\end{equation}
In practice, the prior $p_\theta(z)$ is commonly set as the standard multivariate normal distribution $\mathcal{N}(0,\mathcal{I})$. In those cases, all the features are uncorrelated, and the $KL$ divergence regularizes the latent features distribution $q_\phi(z|\bx^{(i)}$ to a less correlated one. Note that the larger the $\beta$, the less correlated (more disentangled) the features will be.

\section{Applications of autoencoders}
\label{sec:Applications_of_autoencoders}
Learning a representation via the autoencoder can be used for various applications. The different types of autoencoders may be modified or combined to form new models for various applications. For example, in \cite{VariationalAutoEncoderClassification}, they are used  for classification, captioning, and unsupervised learning.
We describe below some of the applications of autoencoders.

\subsection{Autoencoders as a generative model}

\begin{figure}
\centering
\begin{subfigure}{.5\textwidth}
    \centering
  \includegraphics[scale=0.3]{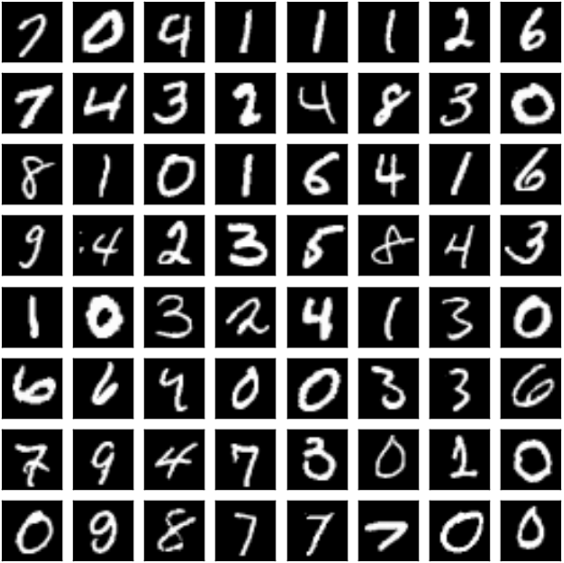}
  \caption{Sample from the original MNIST dataset.}
  \label{subfig:original_MNIST}
\end{subfigure}%
\begin{subfigure}{.5\textwidth}
    \centering
  \includegraphics[scale=0.3]{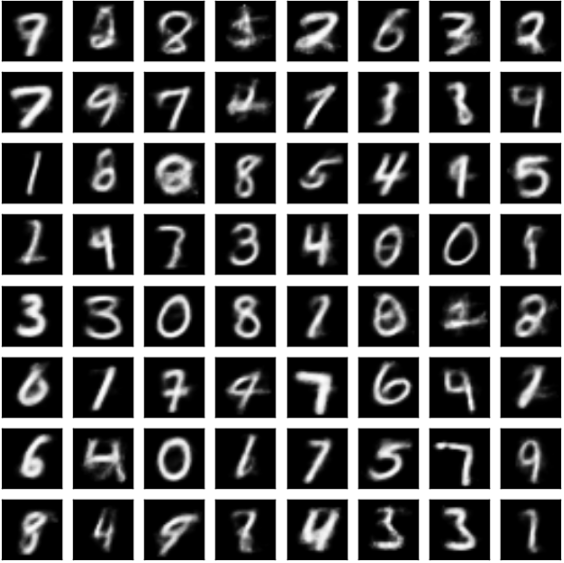}
  \caption{VAE generated MNIST images.}
  \label{subfig:ae_generated_MNIST}
\end{subfigure}
\caption{Generated images of from a variational autoencoder, trained on the MNIST dataset with a prior $p_\btheta(\bz)=\mathcal{N}(0,\mathcal{I})$. Left: original images from the dataset. Right: generated images.}
\label{fig:vae_generation_MNIST}
\end{figure}

As explained in Section~\ref{sec:variational_autoencoders}, variational autoencoders are generative models that attempt to describe data generation through a probabilistic distribution. Furthermore, as can be seen in Equation~\ref{eq:variational_lower_bound}, the posterior distribution $q_\phi(\bz|\bx^{(i)}$ which is derived by the encoder, is regularized towards a continuous and complete distribution in the shape of the predefined prior of the latent variables $p_\btheta(\bz)$. Once trained, one can simply samples random variables from the the same prior, and feed it to the decoder. Since the decoder was trained generate $\bx$ from $p_\btheta(\bx_i|\bz)$, it would generate a meaningful newly-generated sample. In figure \ref{fig:vae_generation_MNIST}, original and generated images are displayed over the MNIST dataset.
When discussing the generation of new samples, the immediate debate involves the comparison between VAE and GANs. An overview on this can be found at Section~\ref{sec:autoencoders_and_generative_adversarial_networks}, and two methods that combine both models can be found at Sections~\ref{sec:ALI} and ~\ref{sec:WAE}.

\subsection{Use of autoencoders for classification}

While autoencoders are being trained in an unsupervised manner (i.e., in the absence of labels), they can be used also in the semi-supervised setting (where part of the data do have labels) for improving classification results. 
In this case, the encoder is used as a feature extractor and is "plugged" into a classification network. This is mainly done in the semi-supervised learning setup, where a large dataset is given for a supervised learning task, but only a small portion of it is labeled. 

The key assumption is that samples with the same label should correspond to some latent presentation, which can be approximated by the latent layer of autoencoders. First, the autoencoders are trained in an unsupervised way, as described in previous sections. Then (or in parallel), the decoder is put aside, and the encoder is used as the first part of a classification model. Its weights may be fine tuned \cite{Why_Does_Unsupervised} or stay fixed during training. A simpler strategy can be found in \cite{autoencoders_classification}, where a support vector machine (SVM) is trained on the output features of the encoder.
In cases where the domain is high dimensional, and the layer-by-layer training is unfeasable, one solution is to train each layer as a linear layer before adding the non linearity. In this case, even with denoising the inputs, there exists a closed form solution for each layer, and no iterative process is needed \cite{Marginalized}.

\begin{figure}
    \centering
    \includegraphics[scale=0.5]{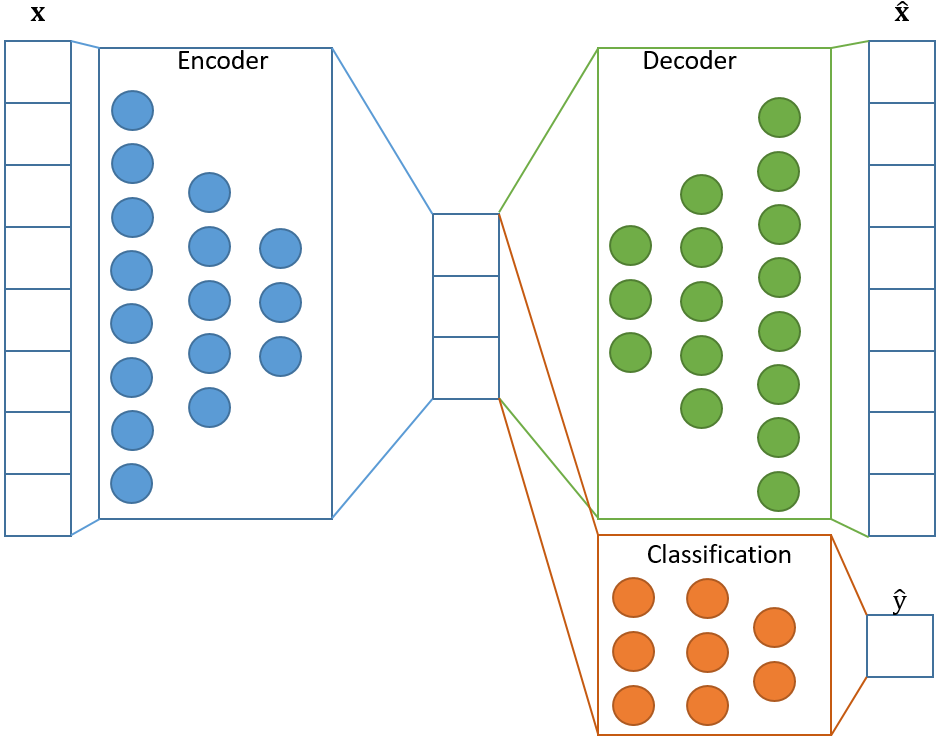}
    \caption{An illustration for using autoencoders as regularization for supervised models. Given the reconstruction loss $R(x,\hat{x})$, and the classification lost function $\mathcal{L}(y,\hat{y})$, the new loss function would be $\Tilde{\mathcal{L}}=\mathcal{L}(y,\hat{y})+\lambda R(x,\hat{x})$, where $\lambda$ is the regularization parameter}.
    \label{fig:supervised_autoencoders}
\end{figure}

Another approach use autoencoders as a regularization technique for a classification network.
For example, in \cite{supervised_autoencoders,Augmenting_Supervised_Neural_Networks}, two networks are connected to the encoder, a classification network (trained with the labelled data) and the decoder network (trained to reconstruct the data, whether labeled or unlabeled). Having the reconstruction head in addition to the classification head serves a regularizer for the latter.
An illustration is given in figure \ref{fig:supervised_autoencoders}.

\subsection{Use of autoencoders for clustering}
Clustering is an usupervised problem, where the target is to split the data to groups such that sampless in each group are similar to one another, and different from the samples in the other groups. Most of the clustering algorithms are sensitive to the dimensions of the data, and suffer from the curse of dimensionality. 

Assuming that the data have some low-dimensional latent representation, one may use autoencoders to calculate such representations for the data,  which are composed of much less features. First, the autoencoder is trained as described in the sections before. Then, the decoder is put aside, similarly to the usage in classification. The latent representation (the encoders output) of each data point is then kept, and serves as the input for any given clustering algorithm (e.g., $K$-means).

The main disadvantage of using vanilla autoencoders for clustering is that the embeddings are trained solely for reconstruction and not for the clustering application. To overcome this, several modifications can be made.
In \cite{Auto-encoder_Based_Data_Clustering}, the clustering is done similarly to the K-means algorithm \cite{kmeans}, but the embeddings are also retrained at each iteration. In this training an argument is added to the autoencoder loss function, which penalizes the distance between the embedding and the cluster center.

In \cite{soft_clustering}, A prior distribution is made on the embeddings. Then, the optimization is done both by the reconstruction error and by the KL-Divergence between the resulting embeddings distribution and the assumed prior.
This can be done implicitly, by training a VAE with the assumed prior. At \cite{autoencoder_gmm}, this is done while assuming a multivariate Gaussian mixture.

\subsection{Use of autoencoders for anomaly detection}
Anomaly detection is another unsupervised task, where the objective is to learn a normal profile given only the normal data examples and then identify the samples not conforming to the normal profile as anomalies.
This can be applied in different applications such as fraud detection, system monitoring, etc.
The use of autoencoders for this tasks, follows the assumption that a trained autoencoder would learn the latent subspace of normal samples. Once trained, it would result with a low reconstruction error for normal samples, and high reconstruction error for anomalies \cite{anomaly_detection1, anomaly_detection2, anomaly_detection3, anomaly_detection4}. 

\subsection{Use of autoencoders for recommendation systems}
A recommender system, is a model or system that seek to predict users preferences or affinities to items \cite{recsys_book}. Recommender systems are prominent in e-commerce websites, application stores, online content providers and have many other commercial applications.
A classical approach in recommender system models is Collaborative Filtering (CF) \cite{CF_Explained}. In CF, user preferences are inferred based on information from other user preferences. The hidden assumption is that the human preferences are highly correlated i.e., people that exhibit similar preferences in the past will exhibit similar preferences in the future. 

An basic example of the use of autoencoders for recommender systems is the AutoRec model \cite{AutoRec}. The AutoRec model has two variants: user-based AutoRec (U-AutoRec) and item-based AutoRec (I-AutoRec). In U-AutoRec the autoencoder learns  a lower dimensional representation of item preferences for specific users while in I-AutoRec, the autoencoder learns a lower dimensional representation of user preferences for specific items. 

For example, assume a dataset consisting of $M$ user and $N$ items. Let $\bf{r}_m \in \mathcal{R}^N$ be a preference vector for the user $m$ consisting of its preference score to each of the $N$ items. U-AutoReco's decoder is $z= g(\mb{r}_m)$ mapping $\mb{r}_m$ into representation the representation vector $z \in \mathcal{R}^{d}$, where $d \ll N$. The reconstruction given the encoder $f(z)$ is $h(\mb{r}_m; \theta)=f(g( \mb{r}_m ))$, where $\theta$ are the model's parameters. 
The U-AutoRec objective is defined as
\begin{equation}
    \argmin_{\theta} \sum_{m=1}^M \| \mb{r}_m -  h( \mb{r}_m; \theta)\|_{O}^2 + \lambda \cdot reg.
\end{equation}
Here, $\| \cdot \|_{O}^2$ means that the loss is defined only on the observed preferences of the user.
At prediction time, we can investigate the reconstruction vector and find items that the user is likely to prefer. 

The I-AutoRec is defined symetrically as follows: Let $\mb{r}_n$ be item $n$'s preference vector for each user.  The I-AutoRec objective is defined as
\begin{equation}
    \argmin_{\theta} \sum_{n=1}^N \| \mb{r}_n -  h( \mb{r}_n; \theta)\|_{O}^2 + \lambda \cdot reg.
\end{equation}
At prediction time, we reconstruct the preference vector for each item, and look for potential users with high predicted preference. 

In \cite{Hybrid_AutoRec1,Hybrid_AutoRec2}, the basic AutoRec model was extended by including de-noising techniques and incorporating users and items side information such as user demographics or item descriptoin. The de-noising serve as another type of regularization that prevent the auto-encoder overfitting rare patterns that do not concur with general user preferences. The side information whas shown to improve accuracy and speed-up the training process.

Similar to the original AutoRec, two symetrical models have been proposed, one that works with user preference $\mb{r}_m$ vectors and the other with item preference vectors $\mb{r}_n$. In the general case, these vectors may consist of explicit ratings. The Collaborative Denoising Auto-Encoder (CDAE) model \cite{Hybrid_AutoRec_implicit} is essentially applying the same approach on vectors of implicit ratings rather than explicit ratings. Finally, a variational approach have been attempted by applaying VAE in a similar fashion \cite{RecSys_VAE}.

\subsection{Use of autoencoders for dimensionality reduction}
Real world data such as text or images is often represented using a sparse high-dimensional representation. While many models and applications work directly in the high dimensional space, this often leads to the \emph{curse of dimensioanlity} \cite{Curse_of_dim}. The goal of dimensionality reduction is to learn a a lower dimensional manifold, so-called ``intrinsic dimensionality'' space. 

A classical approach for dimensionality reduction is Principal Component Analysis (PCA) \cite{PCA}. PCA is a linear projection of data points into a lower dimensional space such that the squared reconstruction loss is minimized. As a linear projection, PCA is optmial. However, non-linear methods such as autoencoders, may and often do achieve superior results. 

Other methods for dimensionalty reduction employ different objectives. For example, Linear Discriminant Analysis (LDA) is a supervised method to find a linear subspace, which is optimal for discriminating data from different classes \cite{duda2001pattern}. ISOMAP \cite{ISOMAP} learns a low dimensional manifold by retaining the geodesic distance between pairwise data in the original space. For a survey of different dimensioanlity methods see \cite{dimensionality_reduction}. 

The use of autoencoders for dimensionality reduction is stright forward. In fact, the dimensionality reduction is performed by every autoencoder in the bottleneck layer. The projection of the original input into the lower-dimensional bottleneck representation is a dimension reduction operation through the encoder and under the objective given to the decoder. For example, an autoencoder comrised of a simple fully connected encoder and decoder with a squared loss objective performs dimension reduction with a similar objective to PCA. However, the non-linearity activation functions often allows for a superior reconstruction when compared to simple PCA. More complex architectures and different objectives allow different complex dimension reduction models.  
To review the different applications of autoencoders for dimension reduction, we erefer the interested reader to \cite{hinton2006reducing,wang2014generalized,wang2016auto}.

\section{Advanced autoencoder techniques}
\label{sec:Advanced_autoencoder_techniques}
Autoencoders are usually trained by a loss function corresponding to the difference between the input and the output. As shown above, one of the strengths of autoencoders is the ability to use their latent representation for different usages. On the other hand, by looking at the reconstruction quality of autoencoders for images, one of its major weaknesses becomes clear, as the resulting images are usually blurry. The reason for that is the used loss function, which does not take into account how realistic its results are and does not use the prior knowledge that the input images are not blurred. In recent years, there were some developments related to autoencoders, which deal with this weakness.

\subsection{Autoencoders and generative adversarial networks}
\label{sec:autoencoders_and_generative_adversarial_networks}
Variational autoencoders are trained (usually) on MSE which yields slightly blurred images, but allows inference over the latent variables in order to control the output.
An alternative generative model to autoencoders that synthesize data (such as images) is the Generative Adversarial Networks (GANs). In a nutshell, a GAN architecture consists of two parts: The generator  which generates new samples, and a discriminator which is trained to distinguish between real samples, and generated ones. The generator and the discriminator are trained together using a loss function that enforces them to compete with each other, and by thus improves the quality of the generated data. This leads to generated results that are quite compelling visually, but in the cost of the control on the resulting images.
Different works have been done for having the advantages of both models, by different combinations of the architectures and the loss functions.
In Adversarial Autoencoders \cite{Adversarial_Autoencoders}, The KL-divergence in the VAE loss function is replaced by a discriminator network that distinguishes between the prior and the approximated posterior. In \cite{decoder_generator}, the reconstruction loss in the VAE loss is replaced by a discriminator, which makes the decoder to essentially merge with the generator. In \cite{GAEL}, the discriminator of the GAN is combined with an encoder via shared weights, which enables the latent space to be conveniently modeled by GMM for inference. This approach was then used in \cite{StyleGANEncoder} for self-supervised learning. 
We detail next two other directions for combining GANs with autoencoders.

\subsection{Adversarially learned inference}
\label{sec:ALI}

One of the disadvantages of GANs is mode collapse, which unlike autoencoders, may cause them to represent via the latent space just part of the data (miss some modes in its distribution) and not all of it.  

\begin{figure}
    \centering
    \includegraphics[scale=0.35]{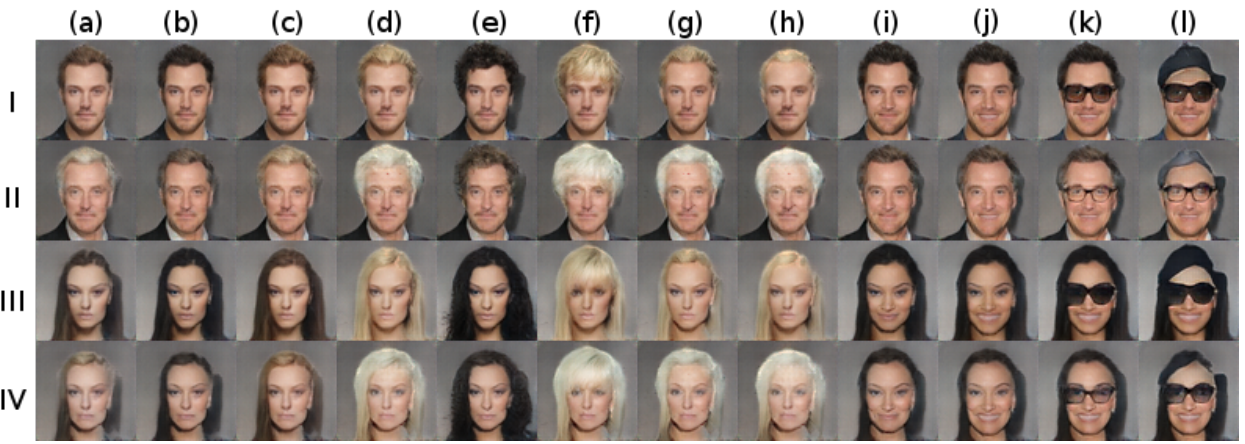}
    \caption{An Image drawn from \cite{ALI}. A model is first trained on the CelebA dataset \cite{CelebA}. It includes 40 different attributes on each image, which in ALI are linearly embedded in the encoder, decoder, and discriminator. Following the training phase, a single fixed latent code $z$ is sampled. Each row has a subset of attributes that are held constant across columns. The attributes are male, attractive, young for row $I$; male attractive, older for row $II$; female, attractive, young for row $III$; female, attractive, older for Row $IV$. Attributes are then varied uniformly over rows across all columns in the following sequence: (b) black hair; (c) brown hair; (d) blond hair; (e) black hair, wavy hair; (f) blond hair, bangs; (g) blond hair, receding hairline; (h) blond hair, balding; (i) black hair, smiling; (j) black hair, smiling, mouth slightly open; (k) black hair, smiling, mouth slightly open, eyeglasses; (l) black hair, smiling, mouth slightly open, eyeglasses, wearing hat.}
    \label{fig:ALI}
\end{figure}

In Adversarially Learned Inference (ALI) there is an attempt to merge the ideas of both VAEs and GANS, and get a compromise of their strengths and weaknesses \cite{ALI}. Instead of training a VAE with some loss function between the input and the output, a discriminator is used to distinguish between $(\bx,\hat{\bz})$ pairs, where $\bx$ is an input sample and $\bz\sim q(\bz|\bx)$ is sampled from the encoders output, and $(\Tilde{\bx},\bz)$ pairs, where $\bz\sim p(\bz)$ is sampled from the used prior in the VAE, and $\Tilde{\bx}\sim p(\bx|\bz)$
is the decoders output.
This way the decoder is enforced to output realistic results in order to "fool" the discriminator. Yet, the autoencoder structure is maintained. An example of how ALI enables altering specific features in order to get meaningful alterations in images is presented in Figure~\ref{fig:ALI}.

ALI is an important milestone in the goal of merging both concepts and it had many extentions.
For example, HALI \cite{HALI} learns the autoencoder in hierchical structure in order to improve the recostruction ability.
ALICE \cite{ALICE} added a conditional entropy loss between the real and the reconstructed images.

\subsection{Wasserstein autoencoders}
\label{sec:WAE}
In continuation to Section~\ref{sec:ALI}, GANs generate compelling images, but do not provide inference, and have a lot of inherent problems regarding its learning stability.
Wasserstein-GAN (WGAN) \cite{WGAN}, solves a lot of those problems by using the Wasserstein distance for the optimizations loss function. The Wasserstein distance, is a specific case of the Optimal Transport distance \cite{OT}, which is a distance between two probabilities, $P_X$ and $P_G$, and is defined as:
\begin{equation}
    W_c(P_X, P_G)=\inf_{\Gamma\in P(X\sim P_X, Y\sim P_G)} \mathbb{E}_{(X,Y)\sim \Gamma}[c(X,Y)]
\end{equation}
where $c(x,y)$ is some cost function. When $c(x,y)=d^p(x,y)$ is a metric measurement, then the $p$-th root of $W_c$ is called the $p$-Wasserstein distance. When $c(x,y)=d(x,y)$, then we get to the 1-Wasserstein distance, which is also known as the "Earth Moving Distance" \cite{EMD} and can be defined as:
\begin{equation}
    W_1(P_X, P_G)=\sup_{f\in\mathfrak{F}}\mathbb{E}_{X\sim P_X}[f(X)]-\mathbb{E}_{Y\sim P_G}[f(Y)]
\end{equation}
Unformally, we try to match the two probabilities by "moving" the first to the latter in the shortest distance, and that distance is defined as the 1-Wasserstein distance.

\begin{figure}
    \centering
    \includegraphics[scale=0.35]{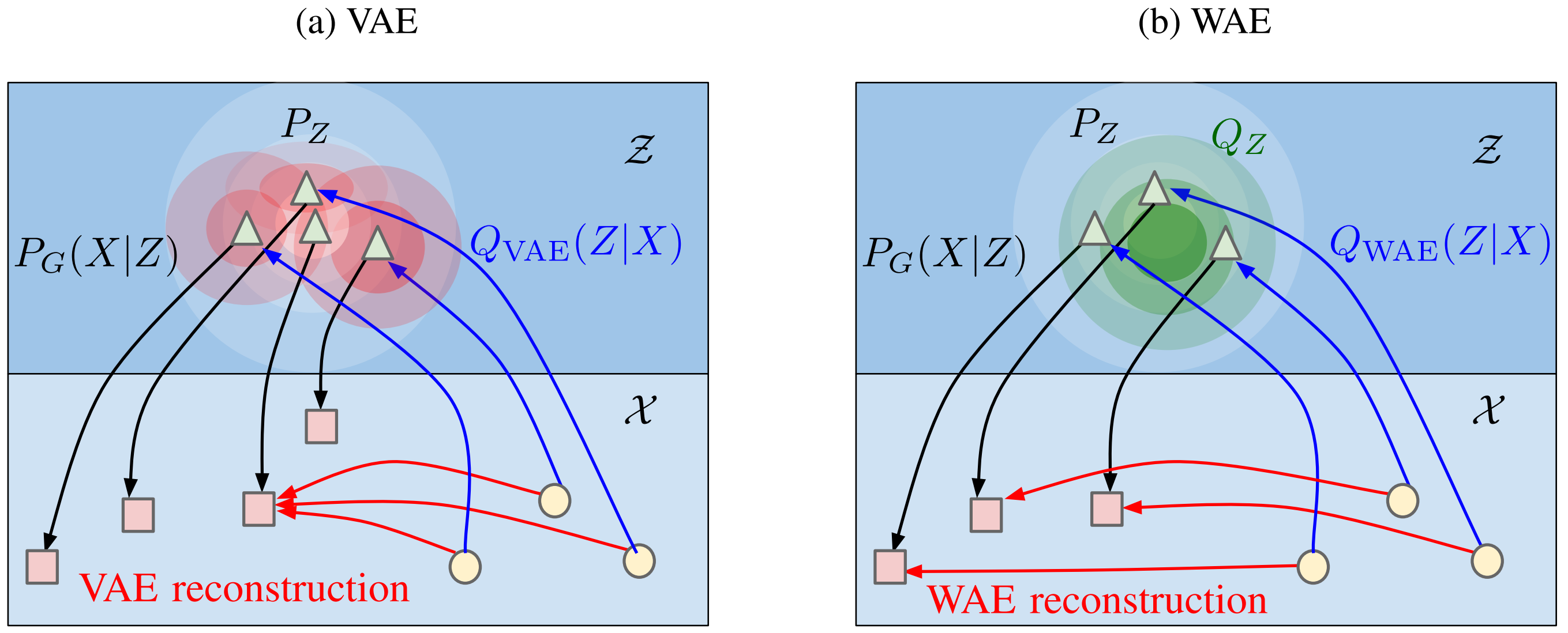}
    \caption{An Image drawn from \cite{Wasserstein_Autoencoders}. Both VAE and WAE minimize two terms: the reconstruction cost and the regularizer penalizing discrepancy between $P_Z$ and distribution induced by the encoder $Q$. VAE forces $Q(Z|X = x)$ to match $P_Z$ for all the different input examples $x$ drawn from $P_X$. This is illustrated on picture (a), where every single red ball is forced to match $P_Z$ depicted as the white shape. Red balls start intersecting, which leads to problems with reconstruction. In contrast, WAE forces the continuous mixture $Q_Z := \int Q(Z|X)dP_X$ to match $P_Z$, as depicted with the green ball in picture (b). As a result latent codes of different examples get a chance to stay far away from each other, promoting a better reconstruction.
}
    \label{fig:WAE}
\end{figure}

As seen in Equation~\ref{eq:variational_lower_bound}, the loss function of a specific sample is comprised of the reconstruction error and a regularization factor which enforces the latent representation to resemble the prior (usually multivariate standard normal). The problem addressed in \cite{Wasserstein_Autoencoders}, is that this regularization essentially pushes all the samples to look the same, and does not use the entire latent space as a whole.
In GANs, the OT distance is used to discriminate between the distribution of real images and the distribution of fake ones. In Wasserstein autoencoders (WAE) \cite{Wasserstein_Autoencoders}, the authors modified the loss function for autoencoders, which lead to the following objective:
\begin{equation}
    D_{WAE}(P_X, P_G)=\inf_{Q(Z|X)\in \mathfrak{Q}} \mathbb{E}_{P_X} \mathbb{E}_{Q(Z|X)}[c(X,G(Z))] +\lambda \cdot  D_Z(Q_Z,P_Z),
\end{equation}
Where $Q$ is the encoder and $G$ is the decoder.
The left part is the new reconstruction loss, which now penalizes on the output distribution and the sample distribution. This is penalization since the "transportation plan" factors through the $G$ mapping \cite{factoring_OT}. The right part penalizes the distance between the latent space distribution to the prior distribution. The authors keep the prior as the multivariate normal distribution, and use to examples for divergences: the Jensen-Shannon divergence $D_js$ \cite{JS}, and the maximum mean discrepancy (MMD) \cite{MMD}
Figure~\ref{fig:WAE} illustrates the regularizations difference between $VAE$ and $WAE$.

\subsection{Deep feature consistent variational autoencoder}
In this section, a different loss function is presented to optimize the autoencoder. Given an original image and a reconstructed one, instead of measuring some norm on the pixel difference (such as the $\ell_2$), a different measure is used that takes into account the correlation between the pixels.

Pretrained classification networks are commonly used for transfer learning. They allow transcending between different input domains, where the weights of the model, which have been trained for one domain, are fine tuned for the new domain in order to adapt to the changes between the domains. This can be done by training all the models' (pretrained) weights for several epochs, or just the final layers. 
Another use  of pretrained networks is style transfer, where a style of one image is transfered to another image \cite{style_transfer}, .e.g., causing a regular photo looks like a painting of a given painter (e.g., Van Gogh) while maintaining its content (e.g., keeping the trees, cars, houses, etc. at the same place). In this case, the pretrained networks serve as a loss function. 

The same can be done for autencoders. A pretrained network can be used for creating a loss function for autoencoders \cite{Deep_feature_VAE}. After encoding and decoding an image, both the original and reconstructed image are inserted as input to a pretrained network. Assuming the pretrained network results with high accuracy, and the domain which it was trained on is not too different than the one of the autoencoder, then each layer can be seen as a successful feature extractor of the input image. Therefore, instead of measuring the difference between the two images directly, it can be measured between their representation in the network layers.
By measuring the difference between the images at different layers in the network imposes a more realistic difference measure for the autoencoder.

\subsection{Conditional image generation with PixelCNN decoders}
\begin{figure}
    \centering
    \includegraphics[scale=0.35]{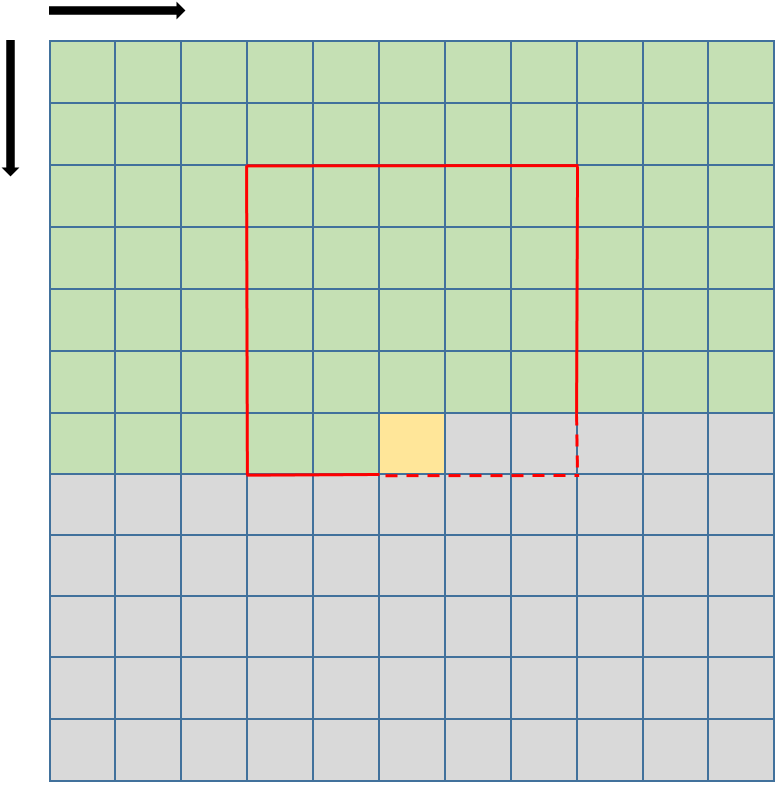}
    \caption{The pixelCNN generation framework. The pixels are generated sequentially. In this case they are generated from top to bottom and from laft to right. The next pixel to be generated is the yellow one. The green pixels are the already generated ones. For generating the yellow pixel, the pixelRNN takes into account the hidden state, and the information of the green pixels in the red square. }
    \label{fig:pixelCNN}
\end{figure}

Another alternative proposes a composition between autoencoders and PixelCNN \cite{pixel_VAE}.
In PixelCNN \cite{pixelCNN}, the pixels in the image are ordered by some arbitrary order (e.g., top to bottom, left to right, or RGB values).
Then the output is formed sequentially where each pixel is a result of both the output of previous pixels, and the input. This strategy takes into account the local spatial statistics of the image, as illustrated in Figure~\ref{fig:pixelCNN}. For example, below a background pixel, there is a higher chance to have another background pixel, than the chance of having a foreground pixel. With the use of the spatial ordering (in addition to the input pixel information), the probability of getting a blurred pixel diminishes. In a later development \cite{pixelRNN}, the local statistics was replaced by the usage of an RNN, but the same concept of pixel generation was remained. This concept can be combined with autoencoders by setting the decoder to be structured as a pixelCNN network generating the output image in a sequential order.

\section{Conclusion}
\label{sec:Conclusions and open problems}
This chapter presented autoencoders showing how the naive architectures that were first defined for them evolved to powerful models with the core abilities to learn a meaningful representation of the input and to model generative processes. These two abilities can be easily transformed to various use-cases,  where part of them were covered.
As explained in Section~\ref{sec:ALI}, one of the autoencoders fall-backs, is that its reconstruction errors do not include how realistic the outputs are.
As for modeling generative processes, despite the success of variational and disentangled autoencoders, the way to choose the size and distribution of the hidden state is still based on experimentation, by considering the reconstruction error, and by varying the hidden state at post training. A future research that better sets these parameters is required.

To conclude, the goal of autoencoders is to get a compressed and meaningful representation. We would like to have a representation that is meaningful to us, and at the same time good for reconstruction. In that trade off, it is important to find the architectures which serves all needs.

\bibliographystyle{spmpsci}
\bibliography{references}

\end{document}